\documentclass[review]{elsarticle}

\usepackage[dvips]{color}
\usepackage{lineno,hyperref}
\journal{Journal of Information Sciences}

\begin{document}

\begin{frontmatter}

\title{Quantifying origin and character of long-range correlations in narrative texts}

\author[IFJ,PK]{Stanis\l{}aw Dro\.zd\.z\corref{mycorrespondingauthor}}
\cortext[mycorrespondingauthor]{Corresponding author}
\ead{stanislaw.drozdz@ifj.edu.pl}

\author[IFJ]{Pawe\l{} O\'swi\c ecimka}
\author[IFJ]{Andrzej Kulig}
\author[IFJ]{Jaros\l{}aw Kwapie\'n}
\author[UJ-F]{Katarzyna Bazarnik}
\author[UJ-Inf]{Iwona Grabska-Gradzi\'nska}
\author[UJ-F]{Jan Rybicki}
\author[PK]{Marek Stanuszek}

\address[IFJ]{Complex Systems Theory Department, Institute of Nuclear Physics, Polish Academy of Sciences, ul.
Radzikowskiego 152, 31-342 Krak\'ow, Poland}
\address[UJ-F]{Institute of English Studies, Faculty of Philology, Jagiellonian University, ul. prof. S. \L{}ojasiewicza
4,
30-348 Krak\'ow, Poland}
\address[UJ-Inf]{Faculty of Physics, Astronomy and Applied Computer Science, Jagiellonian University, ul.\L{}ojasiewicza
11,
30-348 Krak\'ow, Poland}
\address[PK]{Faculty of Physics, Mathematics and Computer Science, Cracow University of Technology, ul. Warszawska 24,
31-155 Krak\'ow, Poland}

\begin{abstract}
In natural language using short sentences is considered efficient for communication. However, a text composed
exclusively  of such sentences looks technical and reads boring. A text composed of long ones, on the other hand,
demands  significantly more effort for comprehension. Studying characteristics of the sentence length variability (SLV)
in a large corpus of world-famous literary texts shows that an appealing and aesthetic optimum  appears somewhere in
between and involves selfsimilar, cascade-like alternation of various lengths sentences. A related quantitative
observation is that the power spectra $S(f)$ of thus characterized SLV universally develop a convincing
`$1/f^{\beta}$' scaling with the average exponent $\beta \approx 1/2$, close to what has been identified
before in musical compositions or in the brain waves. An overwhelming majority of the studied texts simply obeys such
fractal attributes but especially spectacular in this respect are hypertext-like, "stream of consciousness" novels. In
addition, they appear to develop structures characteristic of irreducibly interwoven sets of fractals called
multifractals. Scaling of $S(f)$ in the present context implies existence of the long-range correlations in texts
and appearance of multifractality indicates that they carry even a nonlinear component. A distinct role of the full
stops in inducing the long-range correlations in texts is evidenced by the fact that the above quantitative
characteristics on the long-range correlations manifest themselves in variation of the full stops recurrence
times along texts, thus in SLV, but to a much lesser degree in the recurrence times of the most frequent words.
In this latter case the nonlinear correlations, thus multifractality, disappear even completely for all the texts
considered. Treated as one extra word, the full stops at the same time appear to obey the Zipfian rank-frequency
distribution, however.

\end{abstract}

\begin{keyword}
natural language\sep consciousness\sep correlations\sep multifractals\sep hypertext
\end{keyword}

\end{frontmatter}

%\linenumbers

\section{Introduction}

Mirroring cultural progress~\cite{christiansen2003,agmon2013} during their evolution natural languages - the most
imaginative carriers of information, and the principal clue to the mind and to consciousness~\cite{greenfield2000} -
developed remarkable quantifiable patterns of behaviour such as hierarchical structure in their syntactic
organization~\cite{komarova2001,nowak2002}, a corresponding lack of characteristic
scale~\cite{mitzenmacher2004,newman2005} as evidenced by the celebrated Zipf law~\cite{zipf1949}, small world
properties~\cite{ferrer2001,cong2014,kulig2015}, long-range correlations in the use of
words~\cite{montemurro2002,ausloos2012a,ausloos2012b,altmann2012} or a stretched exponential distribution~\cite{laherrere1998}
of word recurrence times~\cite{altmann2009}. A majority of such patterns are common to a large class of natural
systems known as complex systems~\cite{sole2010,kwapien2012}. With no doubt, language constitutes a great complexity as it
for language is especially true~\cite{hrebicek1999} that "more is different"~\cite{anderson1972} and the capacity of
language
is to generate an infinite range of expressions from the finite set of elements~\cite{hauser2002,ferrer2005}.
Thus this suggests to inspect correlations also among the linguistic constructs longer than mere words.
The most natural of them are sentences - strings of words structured according to syntactical
principles~\cite{akmajian2001,bargiela2006}. Typically it is within a sentence
that words acquire a specific meaning. Furthermore, in a text the sentence structure is expected to be correlated with
the surrounding sentences as dictated by the intended information to be encoded, fluency, rhythm, harmony, intonation
and possibly due to many other factors and feedbacks including the authors' preferences. Consequently, this may
introduce even more
complex correlations than those identified so far. In fact already the Hurst exponent based study~\cite{montemurro2002}
of correlations among words in Shakespeare's plays and in Dickens' and Darwin's books suggests that the range of such
correlations extends far beyond the span of sentences. Indeed, shuffling sentences by preserving their internal structure
appears to bring the Hurst exponents to values even closer to the noise level as compared to their original values. This thus
indicates that long-range correlations among words are induced by factors other than grammar as its range is restricted
essentially to a sentence. As an indication for potential factors generating correlations far beyond the range of single
sentences one should notice that the composition of sentences of varied
length dictates the reading rhythm which involves sound and perception. This, therefore, opens up a possibility that the
Weber-Fechner law~\cite{coren2004} - stating that in perception it is the relative proportions that matter primarily,
and not differences in absolute magnitudes - leaves its imprints also in the sentence arrangement by making some variant
of the multiplicative cascade a likely component of the mechanism that amplifies the associative turns and thus induces
correlations of significantly longer range than the ones due to grammar. There are, however, also 'coarse graining'
constraints to such a mechanism as sentences cannot usually be expanded continuously but by adding clauses, so that
syntactical rules are obeyed. The multifractal formalism~\cite{halsey1986,paladin1987} offers a particularly appropriate
framework to get insight into such effects and to quantify their relative significance and extent.

\section{Materials and Methods}

In order to study the long-range correlations among sentences, particularly those that refer to fractals and cascade
effects, we select a corpus of 113 English, French, German, Italian, Polish, Russian, and Spanish
literary texts of considerable size and for each individually form a series $l(j)$ from the lengths of the consecutive
sentences $j$ expressed in terms of the number of words. Thus, a sentence is defined in purely orthographic terms, as a
sequence of words starting with a capital letter and ending in a full stop. Equivalently, in a text, such a series can be
considered a sequence of the recurrence times of the full stops.
Based on this criterion an initial selection of sentences is performed automatically but then a further
processing is executed, in some cases even manually, in order to identify such (not very frequent) instances where a full
stop does not terminate a sentence, like for instance Mr., in initials, question or exclamation  marks in parenthesis,
etc.
Since the present study has a statistical character, an additional criterion we impose specifies that each text contains
no fewer than 5000 sentences.
For the correlated series, as the ones to be studied here, such a lower bound on the number of sentences is dictated
by requirements to obtain reliable results even on the level of the multifractal analysis~\cite{drozdz2009}.
A complete list of the titles included in this corpus is given in the Appendix.

The simplest, second-order linear characteristics are measured in terms of the power spectra $S(f)$ of such series. Such
spectra are calculated as Fourier Transform modulus squared
\begin{equation}
S(f)=|\sum^{j_{\mathrm{max}}}_{j=1}l(j)e^{-2\pi ifj}|^2
\label{ps}
\end{equation}
of the series $l(j)$ representing lengths of the consecutive sentences $j$.
A complementary approach towards higher order correlations consists in the wavelet decomposition of $l(j)$. The
corresponding `mathematical microscope' wavelet coefficient maps $T_\psi (s,k)$ are obtained as
\begin{equation}
T_\psi (s,k)=(1/\sqrt{s}) \sum^{j_{\mathrm{max}}}_{j=1}l(j)\psi((j-k)/s)
\end{equation}
where $k$ represents the wavelet position in a text while $s$ the wavelet resolution scale.
The wavelet $\psi$ used in the present study is a Gaussian third derivative.
 It is orthogonal - hence insensitive - to quadratic trends in a signal and thus effectively leads to
their
removal~\cite{muzy1994,oswiecimka2006} as demanded by consistency with the other method described and used below.

The wavelet decomposition is optimal for visualization and, in principle, it is well suited to extract the multifractal
characteristics~\cite{muzy1994}. However, the newer method, termed Multifractal Detrended Fluctuation
Analysis (MFDFA)~\cite{kantelhardt2002} is numerically more stable and often more accurate~\cite{oswiecimka2006},
though even here the convergence to a correct result is a delicate matter~\cite{drozdz2009}.
Accordingly, for a series $l(j)$ of sentence lengths one evaluates its signal profile
$L(j) \equiv \sum_{k=1}^j{[l(k)-<l>]}$,
where $<\cdot>$ denotes the series average and $j=1,...,j_{\mathrm{max}}$ with $j_{\mathrm{max}}$ standing for the
number of sentences in a series. This profile is then divided into $2M_s$ disjoint segments $\nu$ of length $s$ starting
from both end points of the series. Next, the detrended variance
\begin{equation}
F^2(\nu,s)=\frac{1}{s} \sum^s_{k=1}\{L((\nu-1)s+k)-P_\nu^{(m)}(k)\}
\end{equation}
is determined, where a polynomial $P_\nu^{(m)}$ of order $m$ serves detrending. Finally, a $q$-th order fluctuation
function
\begin{equation}
F_q(s) = \bigg\{ \frac{1}{2M_s} \sum_{\nu=1}^{2 M_s} [F^2(\nu,s)]^{q/2} \bigg\}^{1/q},
\label{Fq}
\end{equation}
is calculated and its scale $s$ dependence inspected. Scale invariance in a form
\begin{equation}
F_q(s) \sim s^{h(q)}
\label{scaling}
\end{equation}
indicates the most general multifractal structure if the generalised Hurst exponent $h(q)$ is explicitly $q$-dependent,
while it is reduced to monofractal when $h(q)$ becomes $q$-independent. The well-known Hurst exponent is identical to $h(2)$.
$h(q)$ determines the H\"{o}lder exponents $\alpha=h(q)+q h'(q)$ and the singularity spectrum
\begin{equation}
\mathrm{f} (\alpha) =q [\alpha- h (q)] + 1,
\end{equation}
the latter being the fractal dimension
of the set of points with this particular $\alpha$. For a model multifractal series (like a binomial cascade),
$\mathrm{f}(\alpha)$ typically assumes a shape resembling an inverted parabola whose widths $\Delta \alpha =
\alpha_{\mathrm{max}} - \alpha_{\mathrm{min}}$ is considered a measure of the degree of multifractality and thus
often also of complexity.

\section{Results and discussion}

\begin{figure}
\begin{center}
\includegraphics[width=0.9\textwidth]{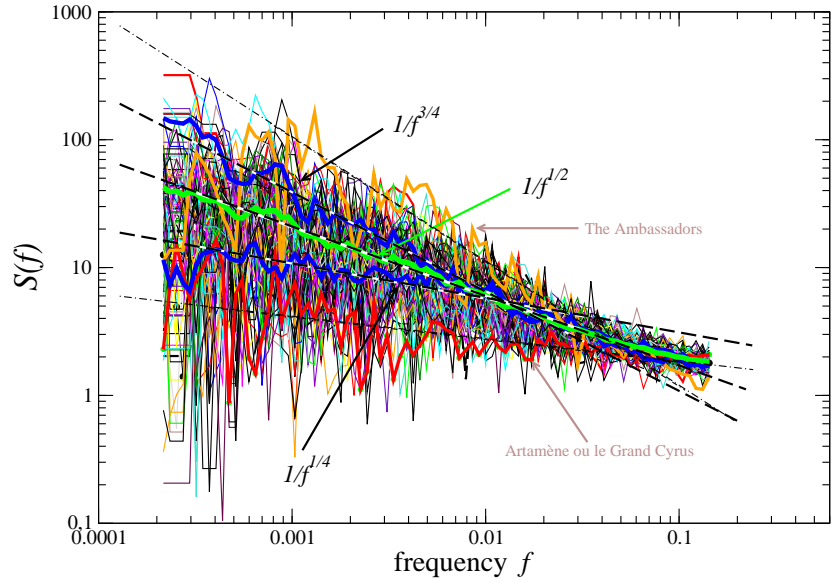}
\end{center}
\caption{Power spectra $S(f)$ of the sentence length variability for 113 world famous literary works. They
are calculated from the series $l(j)$ representing lengths of the consecutive sentences $j$ expressed in terms of the
number of words. $S(f)$ is seen to display $1/f^\beta$ scaling. Middle solid line (green) denotes average over the
individual power spectra, properly normalised, of all the corpus elements and it fits well by $\beta$=1/2. Boundaries of
the dispersion in $\beta$ are indicated by taking average over 10 corpus elements, with the largest $\beta$-values,
which results in $\beta$=3/4 and over 10 its elements with the smallest $\beta$-values, which results in $\beta$=1/4.
The two extremes in the corpus, explicitly indicated, are Henry James's {\it The Ambassadors} (upper) and
Madeleine and/or Georges de Scud\'{e}ry's {\it Artam\`{e}ne ou le Grand Cyrus}, the 17\textsuperscript{th} century novel
sequence (lower), considered the longest novel ever published. The straight line fits to these two extremes are represented
by the dash-dotted lines.}
\label{fig1}
\end{figure}

A highly significant result is obtained already by evaluating the power spectra $S(f)$ according to Eq.(\ref{ps})
of the series representing the sentence length variability (SLV) of all the text considered.
As documented in Fig.~\ref{fig1}, the overall trend of essentially all
sample texts, and especially its average, shows a clear
\begin{equation}
S(f) = 1/f^\beta
\label{sf}
\end{equation}
scaling with $\beta\approx 1/2$, in most cases over
the entire range of about two orders of magnitude in frequencies $f$ spanned by the number of sentences of a typical
text here analyzed. Statistical significance of this result is inspected by randomly shuffling sentences within texts.
For the so obtained randomized texts the calculated power spectra appear to be trivially fluctuating along a horizontal
line ($\beta = 0$) on all the scales.

For the individual original texts $\beta$ is seen to range between 1/4 and 3/4.
This kind of scaling points to the existence of the power-law long-range temporal correlations in
SLV - thus to its fractal organization - and indicates that it balances randomness and orderliness, just as it does
for music, speech~\cite{voss1975}, heart rate~\cite{kobayashi1982},
cognition~\cite{gilden1995}, spontaneous brain activity~\cite{kwapien1998}, and for other `sounds of
Nature'~\cite{bak1996,theunissen2014}. From this perspective human writing appears to correlate with them. Even the
range of the corresponding $\beta$-values from about 1/4 to 3/4 overlaps significantly (more on the Mozart's than
Beethoven's side) with those (1/2 to 1) found~\cite{levitin2012} for musical compositions. This perhaps provides
a quantitative argument for our tendency to refer to writing as `being composed' when we care about all its aspects
including aesthetics and rhythm to be experienced in reading.

The two extremes in the corpus, explicitly indicated in Fig.~\ref{fig1}, are {\it The Ambassadors} (upper)
and {\it Artam\`{e}ne ou le Grand Cyrus}, the 17\textsuperscript{th} century French novel sequence (lower), considered
the longest novel ever published. The first of them appears to be most peculiar among all the texts
as at the small frequencies it visibly departs from $1/f$ by bending down and displaying the two preferred frequencies.
This signals presence of the long-range oscillations with distinct periods in SLV of this particular novel.
At the same time this same novel has the largest $\beta = 0.9 \pm 0.02$ in the region of higher frequencies
where $1/f$ scaling applies.
The latter novel, on the other hand, with $\beta = 0.2 \pm 0.02$ appears closest to the white noise whose $S(f)$ is flat.
It is also appropriate to notice that at the largest frequencies, which corresponds to the smallest scales, the power spectra
of all the texts have some tendency to flatten. This may suggest that the long-range coherence in its $1/f$
organization is somewhat perturbed on shorter scales by coarsening due to grammatical constraints.

\begin{figure}
\begin{center}
\includegraphics[width=0.95\textwidth]{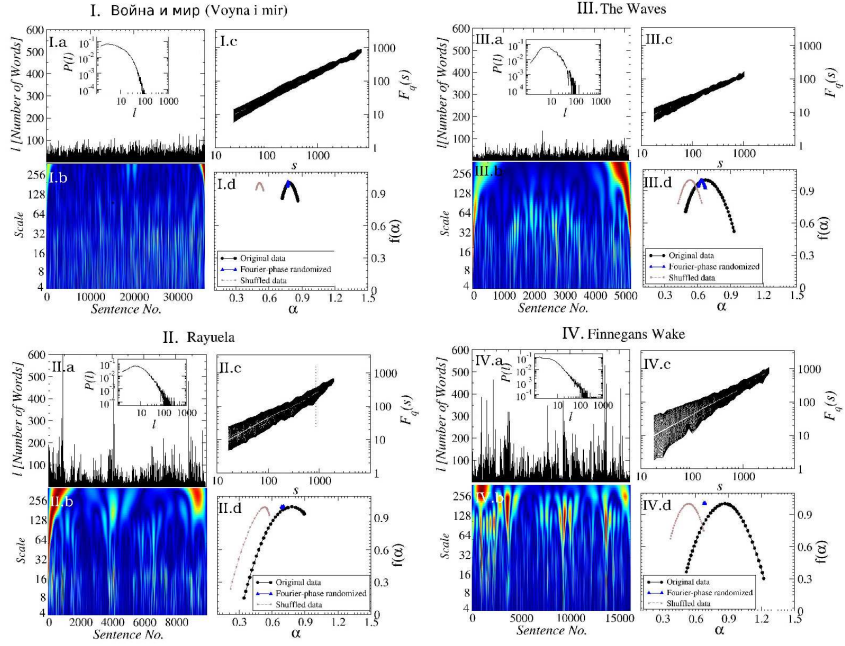}
\end{center}
\caption{Variety of multifractal sentence arrangements in literary texts: Four examples illustrating different
fractal/multifractal characteristics identified within the corpus of the canonical literary texts: {\bf I}, {\it Voyna i
mir (War and Peace)} by Lev Tolstoy; {\bf II}, {\it Rayuela (Hopscotch)} by Julio Cort\'{a}zar; {\bf III}, {\it The
Waves} by Virginia Woolf and {\bf IV}, {\it Finnegans Wake (FW)} by James Joyce. The panels inside each contain
correspondingly: {\bf (a)} The series $l(j)$ of the consecutive sentence lengths throughout the whole text. Insets
illustrate the corresponding probability distributions $P(l)$ of $ l(j)$; {\bf (b)} Wavelet coefficient maps
($T_\psi(s,k)$) obtained for $l(j)$. The wavelet $\psi$ used is a Gaussian third derivative. The horizontal axis
represents the sentence position in a text while the vertical axis - the  wavelet resolution scale $s$. Colour codes
denote magnitude of the coefficient from the smallest (dark blue) to the largest (red); {\bf (c)} $q$-th order
fluctuation functions calculated according to Eq.(\ref{Fq}) using the detrending polynomial $P_\nu ^{(m)}$ of second
order ($m$=2), for $q\in [-4,4]$  and $s\subset [20,j_{max}/5]$ (for {\it Rayuela} a consistent scaling
regime stops somewhat earlier at $s \approx 1000$ which is indicated by the vertical dotted line); 
{\bf (d)} The resulting singularity spectra $\mathrm{f}(\alpha)$ for (i) the series
$l(j)$ representing original texts (black), (ii) for their Fourier-phase randomised counterparts (blue); here
$\mathrm{f}(\alpha)$ is seen shrunk essentially to a point as is characteristic of a pure monofractal, and (iii) for
their randomly shuffled counterparts (gray).}
\label{fig2}
\end{figure}

Our central result relates, however, to the nonlinear characteristics that may manifest themselves in heterogeneous,
self-similarly convoluted structures, undetectable by $S(f)$. Such structures may demand using the whole spectrum of the
scaling exponents and are then termed multifractals. That such structures in SLV may be present within the corpus
analysed here can be inferred from Fig.~\ref{fig2}, which shows four, somewhat distinct, categories of behaviour. A
majority of the texts in our study resembles the case displayed in ({\bf I}). SLV is here seen to be rather homogenously
`erratic' and, consequently, the distribution of cascades seen through the wavelet decomposition is largely uniform. The
three other cases, ({\bf II}), ({\bf III}) and ({\bf IV}), commonly considered representatives of the stream of
consciousness (SoC) literary style that seeks "to depict the multitudinous thoughts and feelings which pass through the mind"
~\cite{cuddon1984}, are visibly inhomogeneous in this respect, as SLV displays clusters of intermittent bursts of much
longer sentences. Such structures are characteristic of multifractals and thus an appropriate subject of the analysis within
the above formalism.

The fluctuation functions $F_q(s)$ obtained according to Eq.(\ref{Fq}) display (Fig.~\ref{fig2}) a convincing scaling
with different degree of $q$-dependence, however. This is corroborated by the corresponding singularity spectra
$\mathrm{f}(\alpha)$, which range from very narrow in {\it Voyna i mir} ({\bf I}), indicating essentially monofractal
structure, through significantly broader - thus already multifractal - but asymmetric, like the strongly left sided {\it
Rayuela} ({\bf II}) or right sided {\it The Waves} ({\bf III}), up to the exceptionally broad and simultaneously almost
symmetric case ({\bf IV}) of {\it Finnegans Wake (FW)}.

\begin{figure}
\begin{center}
\includegraphics[width=0.9\textwidth]{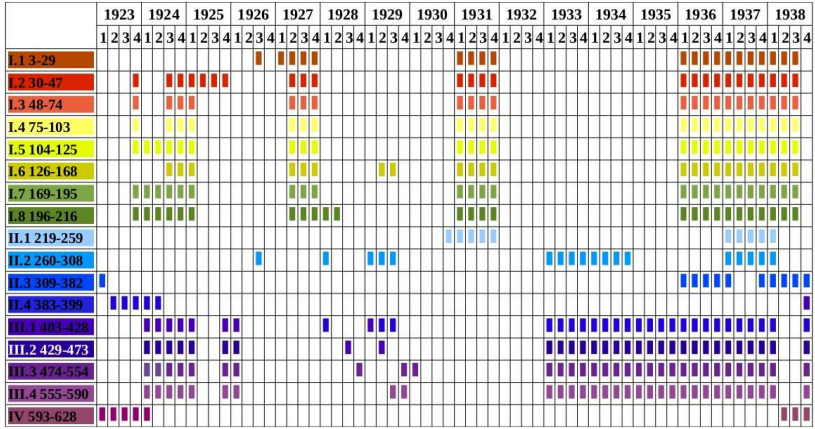}
\end{center}
\caption{Chronological progress of James Joyces engineering work on writing FW, which he described as boring
a mountain from two sides ~\cite{ellmann1982}. This chart may be also taken as a visualisation of Joyces dream about
a Turk picking threads from heaps on his left and right sides, and weaving a fabric in the colours of the
rainbow, which the writer interpreted as a symbolic picture of Books I and III of FW.}
\label{fig3}
\end{figure}

The left side of $\mathrm{f}(\alpha)$ is determined by the positive $q$-values, which filter out larger events (here
longer sentences), and its right side reflects behaviour for smaller events as filtered out by the negative $q$-values.
Hence, asymmetry in $\mathrm{f(}\alpha)$ signals non-uniformity of the underlying hypothesized
cascade~\cite{drozdz2015}. {\it Rayuela} is thus seen to be more multifractal in the composition of long sentences and
almost monofractal on the level of small ones. To some extent the opposite applies to {\it The Waves}. In fact, these
effects can be inferred already from the non-uniformities of the corresponding SLV wavelet decompositions
(Fig.~\ref{fig2}). In this respect {\it FW} appears impressively consistent; being one of the most intriguing literary
`compositions' ever, mastered imaginatively in the SoC technique, freely exploring the mental labyrinth of dreams and
thus often breaking conventional rules of syntax and of linguistic rigour. However, from the perspective of our formal
quantitative approach, its architecture looks - or perhaps just is - a result of these factors - to be governed
consistently by the same `generators' on all scales of sentence length. An extra intellectual factor shaping {\it FW} is
very likely to be also related to its top-bottom development - much like model mathematical cascades - as evidenced by
its chronology of writing~\cite{bazarnik2011} graphically sketched in Fig.~\ref{fig3}.

The significance of the above results for the singularity spectra $\mathrm{f}(\alpha)$ of the series $l(j)$ representing
the original texts has also been tested against the two corresponding surrogates. One standard surrogate in this kind of
analysis is obtained by generating the Fourier-phase randomised counterparts of $l(j)$. This destroys nonlinear
correlations and makes probability distribution of fluctuations Gaussian-like, but preserves the linear correlations
and, as it is clearly seen in Fig.~\ref{fig2}, shrinks $\mathrm{f}(\alpha)$ essentially to a point as is characteristic
of a pure monofractal.  Another surrogate is obtained by randomly shuffling the original series $l(j)$. Consequently,
any temporal correlations get destroyed but the probability distributions of fluctuations remain unchanged. The
corresponding singularity spectra calculated according to the same MFDFA algorithm  are also shown in Fig.~\ref{fig2}
(gray). Consistently with the lack of any temporal correlations they all get shifted down to $\alpha\approx 0.5$ but
some nonzero width of $\mathrm{f}(\alpha)$ still remains to be observed.   However, at least a large part of this
remaining multifractality in this last case may be apparent due to a relatively small size of the samples. For the
uncorrelated series the result of calculating the multifractal spectra is known~\cite{drozdz2009} to end up in either
mono-fractal for the series whose fluctuation probability distributions are L\'{e}vy-unstable, or in bi-fractal for
those whose distributions are L\'{e}vy-stable. Contrary to the correlated series, the convergence to the ultimate
correct results in this case is very slow. We also wish to note at this point that in spite of the Menzerath-Altmann
law, all the relevant results shown here remain essentially unchanged if the sentence length is measured in terms of the
number of characters instead of the number of words.

\begin{figure}
\begin{center}
\includegraphics[width=0.9\textwidth]{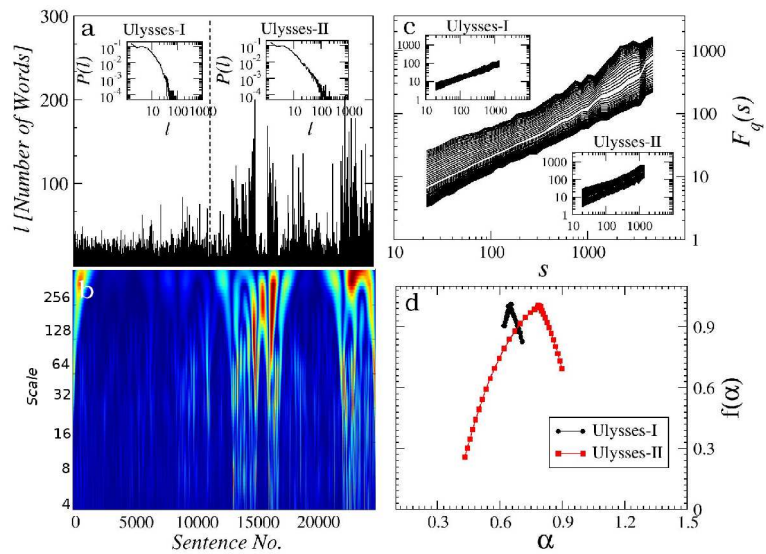}
\end{center}
\caption{Special case of Ulysses: The same convention as in Fig.~\ref{fig2} is used here. The two
additional insets in the panels a and c display results for {\it Ulysses} after bisecting it into halves. {\it Ulysses}-
I corresponds to the text from the beginning to the end of Chapter 10 and {\it Ulysses}-II to the remaining
text without its last two disproportionately long sentences.}
\label{fig4}
\end{figure}

Another, even better known SoC novel by Joyce - {\it Ulysses}, which played a central role in formulating the scale-free
word rank-frequency distribution law by Zipf - also deserves here an extended attention, however, for a
different reason. For this novel, as illustrated in Fig.~\ref{fig4} no unique multifractal scaling can be
attributed, and thus no $\mathrm{f}(\alpha)$ assigned. The SLV inspected both in terms of the sentence length
distribution and through its wavelet transform indicate clearly that {\it Ulysses} splits into two parts such that each
of them may independently have well defined scaling properties. Indeed, by bisecting it approximately into halves
(between Chapters 10 and 11) allows us again to comprise {\it Ulysses} within the present formalism. The first part
appears essentially monofractal, while the other is clearly multifractal, though asymmetrically left-sided, just as {\it
Rayuela}. In fact, this result provides a quantitative argument in favour of the ''doubleness'' of {\it
Ulysses}~\cite{mcHale1992}.

\begin{figure}
\begin{center}
\includegraphics[width=1\textwidth]{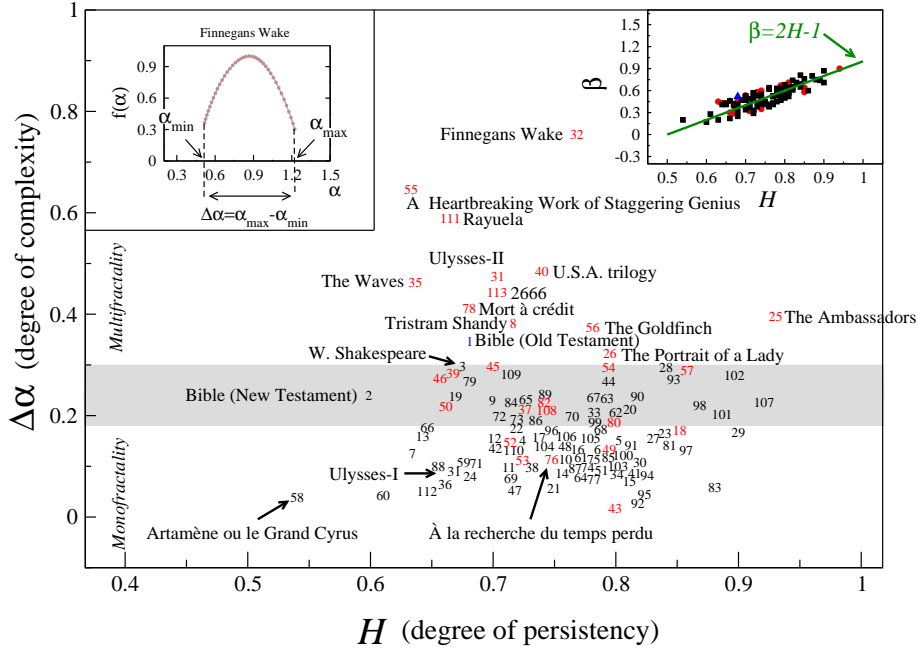}
\end{center}
\caption{ `Scatter plot', which for a collection of 113 most
representative literary works indicated by their numbers  (the ones in red indicate those works that
usually 
are considered as belonging to the SoC narrative) on our list (see Appendix) displays the width $\Delta \alpha$
(schematically defined in the upper-left inset) and the Hurst exponent $H$. Shaded area marks
the transition (uncertainty) region between fully developed multifractality and definite monofractality. We find it
reasonable to assume that the shuffled series are mono-fractal (or at most bi-fractal) and that any trace of
multifractality in this case is an artifact of the finiteness of a series. Therefore, the lower bound of the shaded area
is determined as an average of $\Delta \alpha$'s for all the series (texts) shuffled. Due to the thickest tails in the
probability distributions $P(l)$ of $l(j)$ in {\it FW} (seen in the inset to panel {\bf IV} of Fig.~\ref{fig2}), which
after shuffling the corresponding series may yield the strongest apparent multifractality signal, the upper bound of the
uncertainty region is taken as $\Delta \alpha$ of the shuffled {\it FW}. 
 The upper-right inset shows location of a pair of the $H$ and $\beta$ values for each book, represented by
a point, relative to the straight line determined by Eq.~(\ref{betah}).
}
\label{fig5}
\end{figure}

The results, represented in terms of the width $\Delta \alpha= {\alpha}_{max} - {\alpha}_{min}$ of $\mathrm{f}(\alpha)$,
where $\alpha_{max}$ and $\alpha_{min}$ denote the beginning and the end of $\mathrm{f}(\alpha)$ support,
and of the Hurst exponents $H=h(2)$ measuring the degree of persistence in SLV, for the whole studied corpus are
collected in Fig.~\ref{fig5}. The relation between the Hurst exponent $H$ and the scaling exponent $\beta$
in Eq.~(\ref{sf}) reads~\cite{heneghan2000}:
\begin{equation}
\beta = 2 H - 1.
\label{betah}
\end{equation}
As the upper-right inset to Fig.~\ref{fig5} visibly documents this relation appears to be very
satisfactorily fulfilled when comparing the results presented in Fig.~\ref{fig1} versus the Hurst exponents in
Fig.~\ref{fig5} for all the texts studied which thus provides an independent test for correctness of the results
presented. All the explicit values of $H$ and $\beta$ correspondingly, together with their error bars
$\sigma_H$ and $\sigma_{\beta}$ measured in terms of the mean standard deviations, are listed in the Appendix in
parallel with the titles included in the corpus.

The `scatter plot' shown in Fig.~\ref{fig5} opens up room for many further interesting observations and hypotheses
or even definite conclusions of general interest. Some of them can be straightforwardly listed as follows:
(i) Essentially all the studied texts that are seen in the multifractality region are commonly classified as SoC
literature. The only exception found here, the {\it Old Testament}, has not been considered before in this context.
(ii) $\Delta \alpha$ for all the texts that do not belong to SoC is located below the border of definite
multifractality. Their complexity is thus poorer.
(iii) Also, several texts, by some considered as SoC, appear to be located significantly below this border. An important
example of this is {\it \`{A} la recherche du temps perdu} by Marcel Proust (no. 76 in the list given in the Appendix),
which is clearly monofractal.
(iv) {\it Artam\`{e}ne ou le Grand Cyrus}  is seen to have characteristics
just opposite to {\it FW}. Here $\Delta \alpha$ equals nearly zero and $H$ gets shifted down towards
1/2, which complements its flat power spectrum seen in Fig.~\ref{fig1}, to mean that the corresponding SLV is of the
white noise type.

\begin{figure}
\begin{center}
\includegraphics[width=0.8\textwidth]{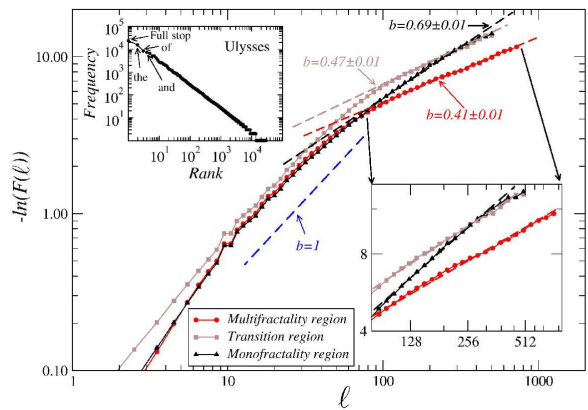}
\end{center}
\caption{Sentence length complementary cumulative distributions
$F(\ell) = Pr(l \ge \ell)$ for the whole corpus of 113 books considered but split into three groups according
to their location in Fig.~\ref{fig5}.
Tails ($\ell > 100$) of these distributions are fitted with the stretched exponential $F(\ell) = \exp{(-\mu
\ell^b)}$ and the corresponding best fit $b$-values listed. The intermediate $(10 < \ell < 100)$ region is seen to
obey a pure exponential $(b = 1)$ dependence. Inset shows the Zipfian rank-frequency distribution plot for {\it Ulysses}
where the full stops are treated as another word.}
\label{fig6}
\end{figure}

Finally, examples of the sentence lengths distributions shown in Fig.~\ref{fig2} ({\bf Ia} to {\bf IVa}) indicate some
differences among works,  related more to the style than to the language.
Those representing SoC kind of narrative have a clear tendency to develop thicker tails. 
 In fact, attempts to exploit the sentence length distribution in stylometry and in authorship attribution 
have a long history~\cite{yule1938,wake1957,sichel1974,ausloos2000}.
In order to approach this issue quantitatively for the present corpus the complementary cumulative distributions 
$F(\ell) = Pr(l \ge \ell)$ for
all the texts collected separately from the three distinct regions identified through Fig.~\ref{fig5} are displayed in
Fig.~\ref{fig6}. The tails of these distributions, starting at around $l \approx 100$, appear to be well represented by
the stretched exponential $F(\ell) = \exp{(-\ell^b)}$ with the smallest $b$ parameter and thus the thickest tail
for the SoC narrative, thus multifractal in SLV. This parameter remains, however, significantly smaller than unity even
in the monofractal regime. This result closely resembles the stretched exponential distribution of recurrence times
between words~\cite{altmann2009}. In a sense the sentence length can be interpreted as a recurrence time between the
full stops and they, in a text, play at least as important role as words. Their large recurrence times appear to be
governed by a similar distribution as the one for words. Interesting in this context, and also complementary,
is observation documented in the upper-left inset of Fig.~\ref{fig6}, that the full stops (dots, question and
exclamation marks) treated as one more word in the rank-frequency dependance typically belong to the same Zipfian plot
as worlds, as shown for the original Zipf's example of {\it Ulysses}.

\begin{figure}
\begin{center}
\includegraphics[width=0.8\textwidth]{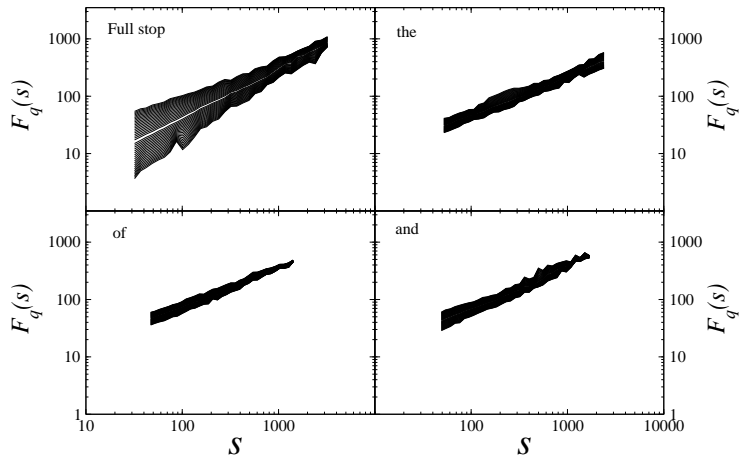}
\end{center}
\caption{ The $q$-th order fluctuation functions calculated using the same algorithm and convention as in
Fig.~\ref{fig2} for the sentence length variability (SLV), which is equivalent to the recurrence times of the full
stops, versus the series of recurrence times of three different words, {\it the}, {\it of} and {\it and} considered
separately, in {\it Finnegans Wake} by James Joyce.}
\label{fig7}
\end{figure}

\begin{figure}
\begin{center}
\includegraphics[width=1\textwidth]{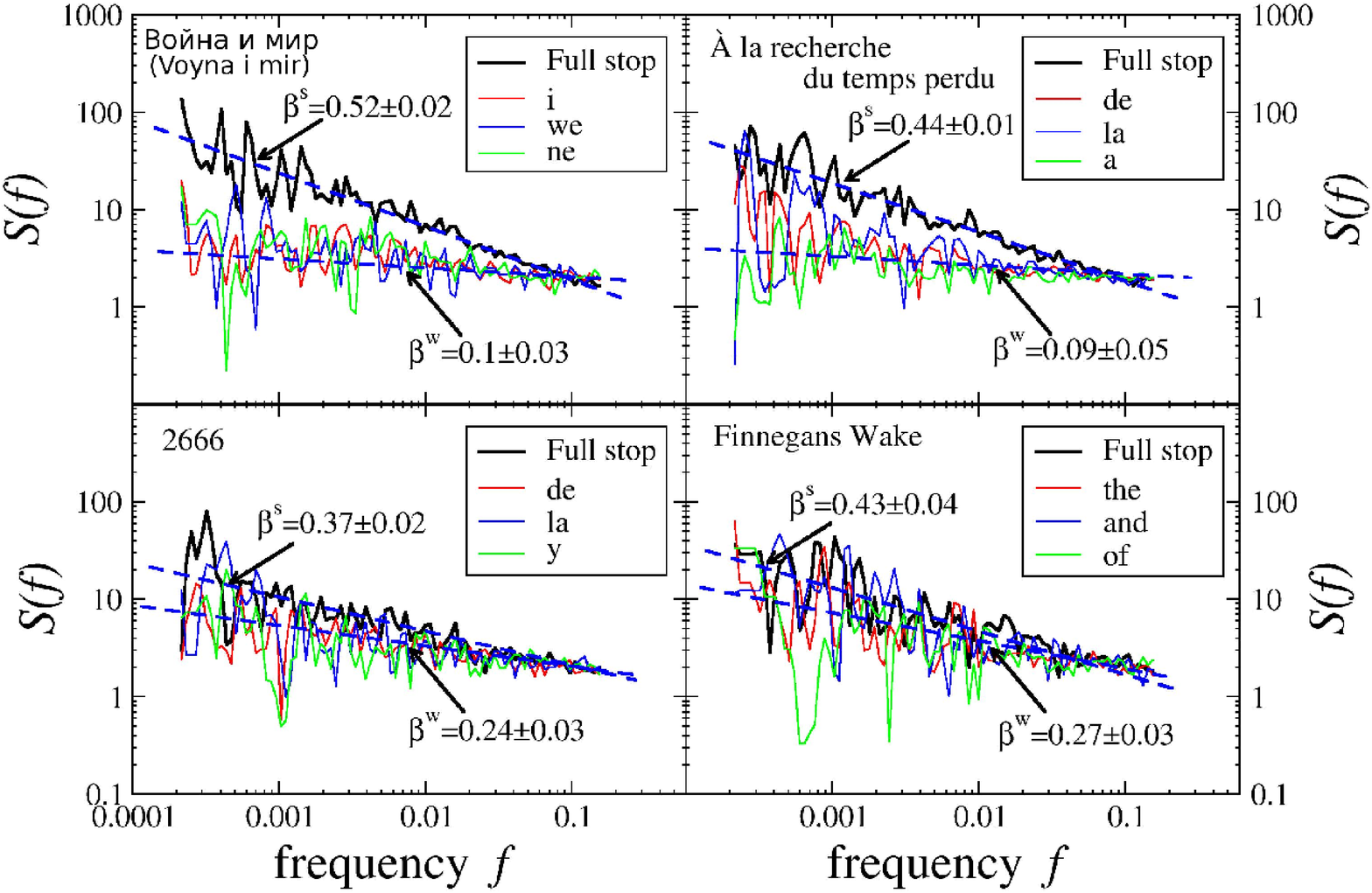}
\end{center}
\caption{ Power spectra $S(f)$ of the sentence length variability (scaling exponent $\beta^s$) and of the series of
recurrence times for the three most frequent words in four texts representative for the present corpus. Straight line
fits are indicated by the dashed lines. The $S(f)$ for the recurrence times is fitted globally and the corresponding
scaling exponent denoted by $\beta^w$.}
\label{fig8}
\end{figure}

These observations thus prompt a question if a series of recurrence times (measured in terms of the number of words)
between the same words develop similar long-range correlations as the ones identified above for the full stops.
For the lowest-rank - thus most frequent - words like English {\it the}, {\it and}, {\it of} or {\it to}
(only for such words one obtains sufficiently long series)
multifractality turns out to disappear in all the texts studied. This is true even for {\it FW} as the $q$-th order
fluctuation functions $F_q(s)$ calculated according to Eq.(\ref{Fq}), for this originally extreme case of
multifractality and displayed in Fig.~\ref{fig7} show. For the series of recurrence times of the words {\it the}, {\it
and} and {\it of} the $F_q(s)$ functions become very weakly $q$-dependent as is characteristic to monofractals. In fact
even this monofractal scaling is here not fully convincing at some places. The long-range nonlinear correlations
responsible for multifractality appear thus to be originating exclusively from the specific arrangement of the full
stops in texts. The situation with the long-range linear correlations is not that extreme as some representative cases
expressed in terms of the power spectra $S(f)$ shown in Fig.~\ref{fig8} illustrate it. These power spectra for other
cases than SLV still show a trace of scaling of the form $S(f) = 1/f^\beta$ with significantly smaller values of the
$\beta$-parameter (denoted by $\beta^s$ for SLV and by $\beta^w$ for the word recurrence times),
however, and larger dispersion of fluctuations along the fit which indicates a very weak character
of the underlying long-range linear correlations. Very interestingly, however, for those texts that are multifractal
on the level of SLV, the linear long-range correlations for the recurrence times between words do not depart so much
from the correlations measured by SLV in the same texts and are significantly stronger than in the monofractal texts,
as the examples in Fig.~\ref{fig8} show.

\section*{Summary}
The present analysis, based on a large corpus which includes world famous literary texts, uncovers the long-range
correlations in their sentence arrangement. The linear component of these correlations universally reveals the
scale-free `$1/f^\beta$' form as characteristic to many other `sounds of Nature' and thus this observation may serve as
an indicator of those factors that shape human language.  The corresponding $\beta$-value typically range from about 1/4
to 3/4 and may thus serve also as a very useful and inspiring stylometry measure. As far as correlations in the sentence
length variability are concerned, some texts - within the present corpus exclusively belonging to the stream of
consciousness narrative - develop even more complex scale-free patterns of the nonlinear character and heavy tailed
intermittent bursts in SLV similar to the ones identified in other areas of human activity~\cite{barabasi2005}. In
quantitative terms this results in a whole spectrum of the scaling exponents as compactly grasped by the multifractal
spectrum $\mathrm{f}(\alpha)$, whose width reflects the degree of nonlinearity involved. A greater complexity of such
hypertext-like narrative finds an interesting parallel in the biological dynamical system as documented~\cite{ivanov1999}
for the healthy human heartbeat, which develops broader multifractal spectra as compared to the heart disfunction. That the
SoC kind of narrative should simultaneously activate greater variety of brain areas seems quite natural. Whether this
indicates means to more efficient sharing of information also emerges as an intriguing perspective to study.
A further argument in favour of such a likely correspondence is that hypertext is paralleled by the underlying
architecture of World Wide Web, which proves easy-to-use and flexible in its self-similar traffic~\cite{crovella1997} of
sharing information over the Internet, indeed.

SLV is equivalent to variability of the recurrence times, measured in terms of the number of words between the full
stops. Analogous variability of the recurrence times between words, which for statistical reasons can be studied for the most
frequent ones, also involves `$1/f^\beta$'-type long-range correlations but they appear significantly weaker than for the full
stops. What is more, the nonlinear correlations inducing multifractal characteristics seem to take place exclusively on the
level of SLV. Thus, even though the full stops together with words appear to belong to the same Zipfian distribution, they form
a frame for long-range correlations in narrative texts. This frame seems to obey more strict universal principles of organization,
likely shaped also by the factors listed in the introduction, and words have some more freedom in filling and
complementing it as can be inferred from their weaker mutual correlations. Such a scenario offers one possible visualization
of the results obtained.

 Finally, the results presented - like imprints of the Weber-Fechner law through SLV cascades - appear
largely
consistent with the working hypothesis formulated in the introduction by listing factors that may induce long-range correlations
during the process of the narrative text formation. In order to further illuminate on such issues some empirical study,
like quantifying the vocal and perception characteristics of texts with varying strength of SLV correlations, would be crucially
helpful and our results indicate direction.

\section*{Acknowledgment}

We thank Krzysztof Bartnicki (who translated {\it FW} into Polish) for constructive exchanges at the
early stage of this Project.

\newpage
\section*{Appendix}

List of the considered literary works.

\begin{figure}[h]
\begin{center}
\includegraphics[width=0.96\textwidth, clip=true]{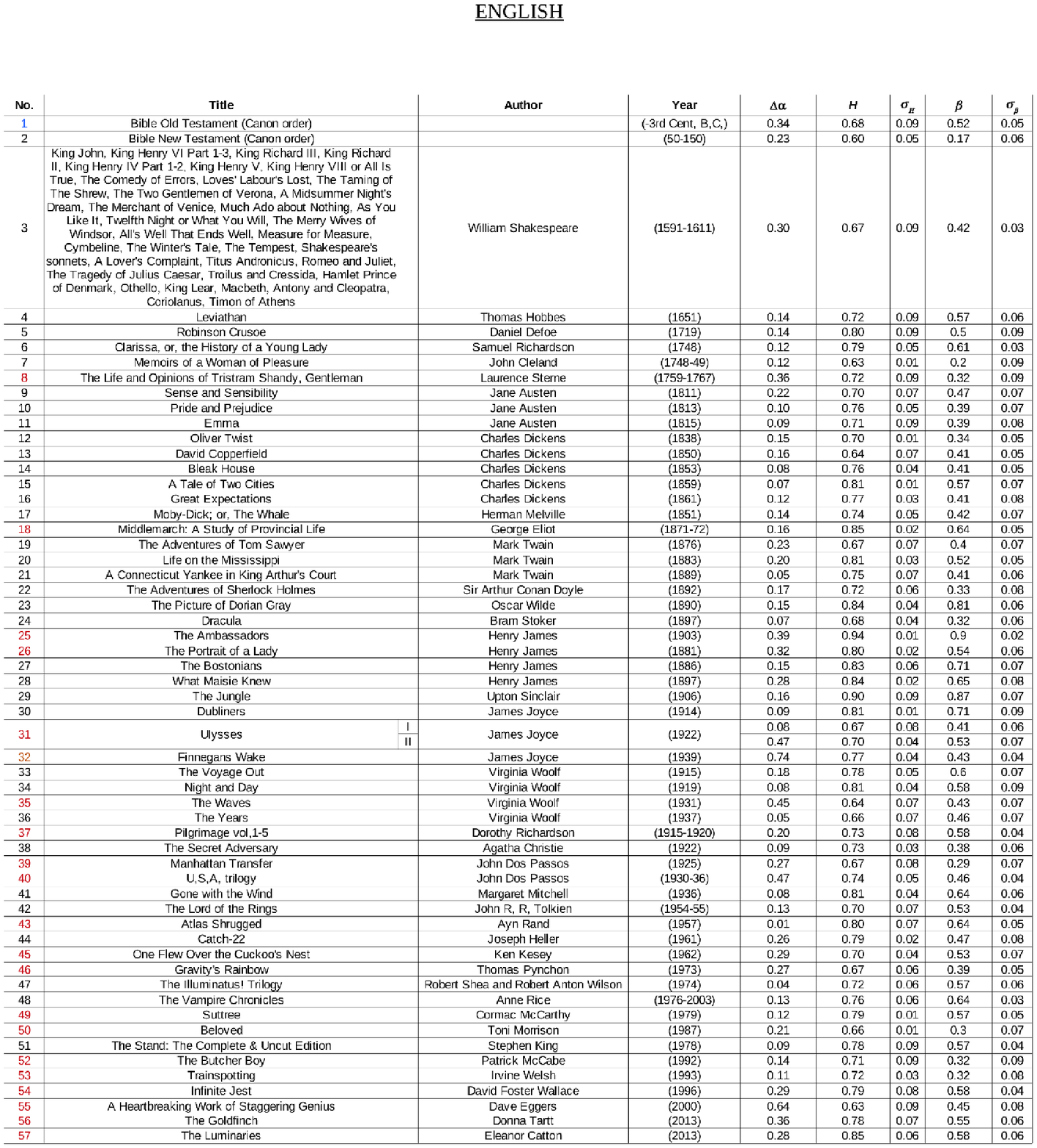}
\end{center}
\label{lista1}
\end{figure}

\newpage
\begin{figure}[h]
\begin{center}
\includegraphics[width=0.96\textwidth, clip=true]{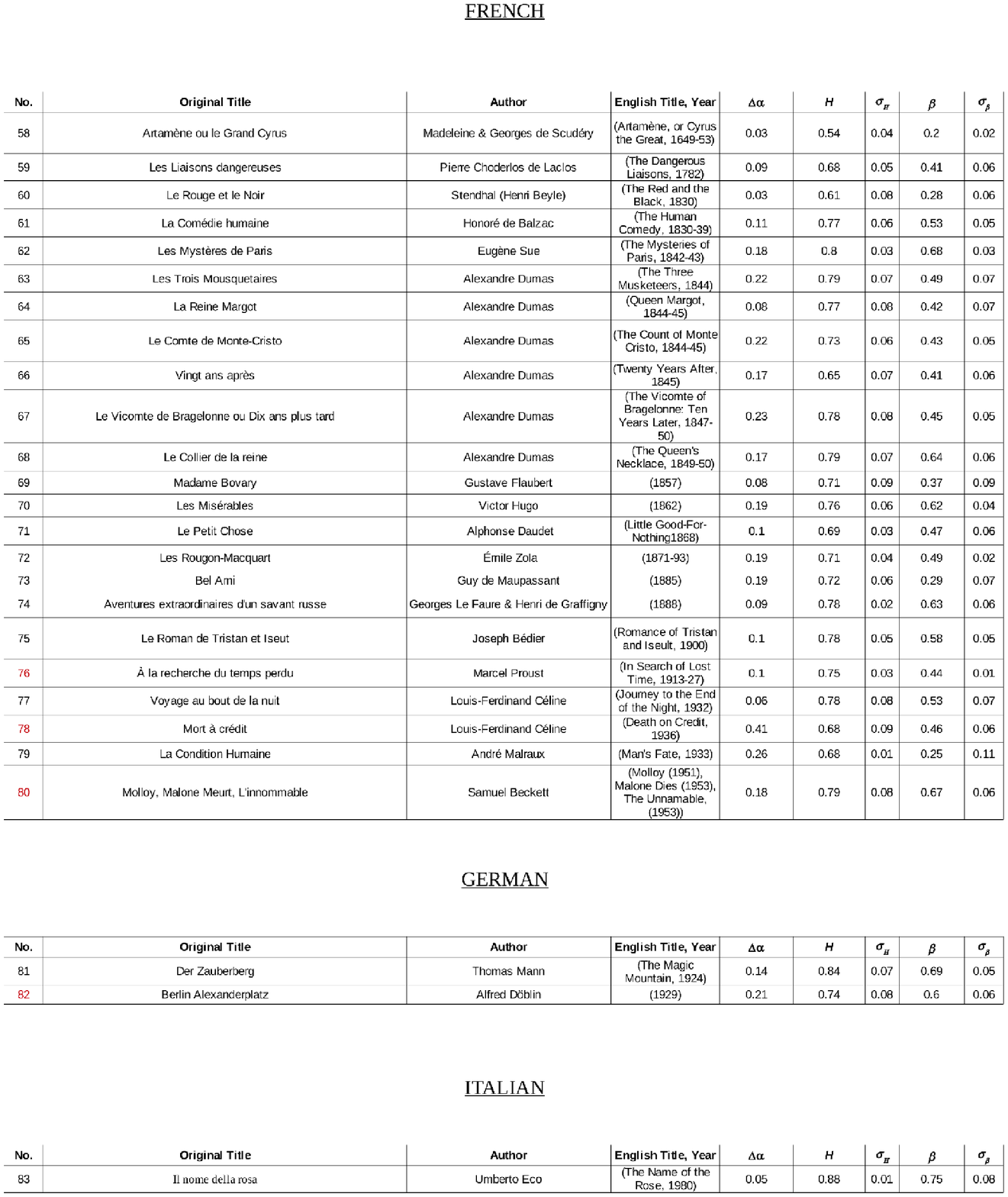}
\end{center}
\label{lista2}

\newpage
\end{figure}
\begin{figure}[h]
\begin{center}
\includegraphics[width=0.96\textwidth, clip=true]{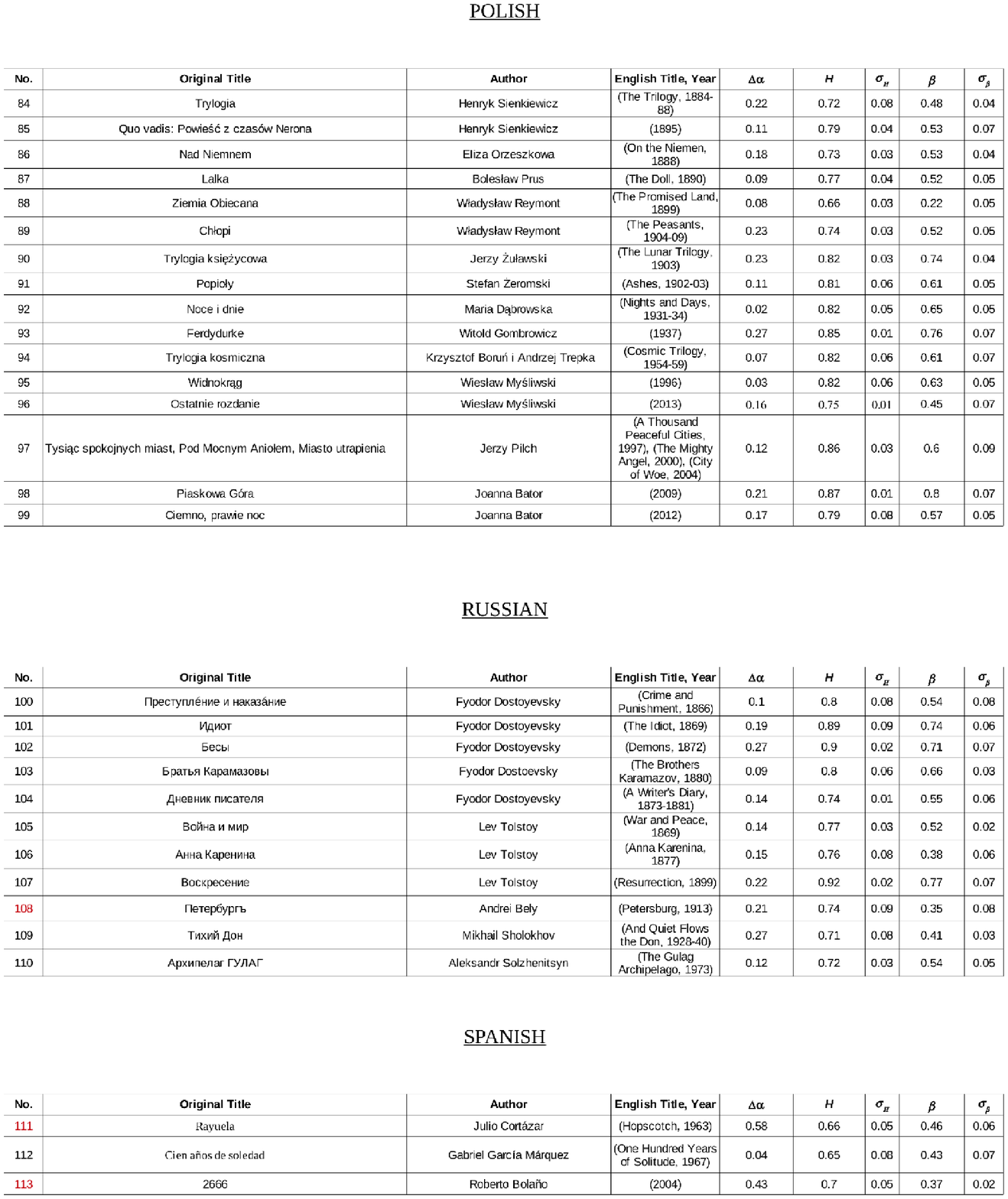}
\end{center}
\label{lista3}
\end{figure}

\end{document}